\documentclass[runningheads]{llncs}
\usepackage{graphicx}
\usepackage{amsmath}
\usepackage{multirow}
\usepackage{amssymb}
\usepackage{hyperref}
\usepackage[misc]{ifsym}

\begin{document}
\title{Discrepancy-based Active Learning for Weakly Supervised Bleeding Segmentation in Wireless Capsule Endoscopy Images}
\titlerunning{DEAL for Weakly Supervised Bleeding Segmentation}

\author{Fan Bai\inst{1} \and
Xiaohan Xing\inst{2} \and
Yutian Shen\inst{1} \and
Han Ma\inst{1} \and
Max Q.-H. Meng\inst{1,3}}
% \author{Anonymous}
%
\authorrunning{Fan Bai, et al.}
% \institute{Anonymous Organization\\
% \email{***@*****.***}}

\institute{Department of Electronic Engineering, The Chinese University of Hong Kong,\\ Shatin, Hong Kong \\
\email{\{fanbai,yt.shen,hanma\}@link.cuhk.edu.hk}
\and
Department of Electrical Engineering, City University of Hong Kong,\\ Kowloon, Hong Kong \\
\email{xiaoxing@cityu.edu.hk}
\and
Department of Electronic and Electrical Engineering, Southern University of Science and Technology, Shenzhen, China\\
\email{max.meng@sustech.edu.cn}
}

\maketitle              % typeset the header of the contribution
\begin{abstract}
Weakly supervised methods, such as class activation maps (CAM) based, have been applied to achieve bleeding segmentation with low annotation efforts in Wireless Capsule Endoscopy (WCE) images. However, the CAM labels tend to be extremely noisy, and there is an irreparable gap between CAM labels and ground truths for medical images. This paper proposes a new Discrepancy-basEd Active Learning (DEAL) approach to bridge the gap between CAMs and ground truths with a few annotations. Specifically, to liberate labor, we design a novel discrepancy decoder model and a CAMPUS (CAM, Pseudo-label and groUnd-truth Selection) criterion to replace the noisy CAMs with accurate model predictions and a few human labels. The discrepancy decoder model is trained with a unique scheme to generate standard, coarse and fine predictions. And the CAMPUS criterion is proposed to predict the gaps between CAMs and ground truths based on model divergence and CAM divergence. We evaluate our method on the WCE dataset and results show that our method outperforms the state-of-the-art active learning methods and reaches comparable performance to those trained with full annotated datasets with only 10\% of the training data labeled.

\keywords{Active Learning \and Segmentation \and WCE Images.}
\end{abstract}
\section{Introduction}
Wireless Capsule Endoscopy (WCE)~\cite{jia2019wireless} is a first-line diagnostic tool for GI tract cancers due to its non-invasiveness to patients. It can capture images of the entire gastrointestinal tract, allowing visualization and diagnosis of the abnormalities and diseases in the GI tract. In recent years, researchers have paid more attention to the problem of abnormality classification~\cite{xing2021categorical,muruganantham2022attention} and detection~\cite{goel2022dilated}, such as bleeding, polyps, inflammatory and other abnormalities in WCE images. Compared to these, abnormality segmentation~\cite{jia2021multibranch} is more challenging due to the complexity of the task and the mass annotation cost caused by pixel-level labels.

Weakly supervised methods, especially CAM-based~\cite{zhou2016learning,selvaraju2017grad}, have attracted increasing attention due to high data efficiency~\cite{wu2019weakly,tang2021m}. Without any pixel-level labels, these methods are capable of achieving segmentation relying on the CAMs. 
However, there is an irreparable gap between the generated CAMs and the ground truths, even though weakly supervised learning is constantly evolving. We experimentally find that existing weakly supervised methods\cite{selvaraju2017grad} perform poorly on the WCE bleeding segmentation task due to the gaps between CAMs and ground truths. Since medical diagnosis requires exceptionally high accuracy, bridging the gap is critical to the practical application of weakly supervised methods. Intuitively, replacing the nasty CAMs with more accurate pseudo labels and ground truths will revitalize these methods from a data perspective and makes the performance of weakly supervised learning infinitely close to that of full supervision. But how to pick out the nasty CAMs is a critical problem.

Active learning is an efficient data selection strategy that selects the most informative samples for annotation based on uncertainty~\cite{caramalau2021gcn}, data distribution~\cite{sener2017coreset,sinha2019vaal}, model gradient~\cite{dai2020ggs}, and other criteria. These methods work well utilizing only unlabeled data. However, in weakly supervised training, all data are annotated by rough CAMs rather than nothing. Therefore, estimating the CAM uncertainty is far more important than traditional criteria under weak supervision. Moreover, previous active learning mainly focused on selecting the human labels, ignoring the accurate pseudo labels from model predictions. However, these accurate pseudo labels can also be selected to replace the nasty CAMs under weak supervision. Since pseudo labels do not increase the labeling burden, it is cost-effective to design a criterion that skillfully combines pseudo labels and ground truths in active learning.

In this paper, we propose the first label selection method to advance weakly supervised to approach fully supervised performance. Our contribution consists of three parts: (1) We design a novel Discrepancy-basEd Active Learning (DEAL) pipeline to select the nasty CAMs and replace them with pseudo labels and a few human annotations, which achieves superior performance; (2) We build a new discrepancy decoder model and design a novel scheme to train it with different propensities to produces standard, coarse and fine predictions, while avoiding unstable single predictions and noisy CAMs; (3) We propose the CAMPUS criterion based on model divergence and CAM divergence, selecting pseudo labels and ground truths to trade off labeling burden and performance. The results show our DEAL outperforms other active learning methods, achieving a comparable performance of full annotated training and saving 90\% human labels.

\begin{figure}[tp!]
\centering
\includegraphics[width=\textwidth]{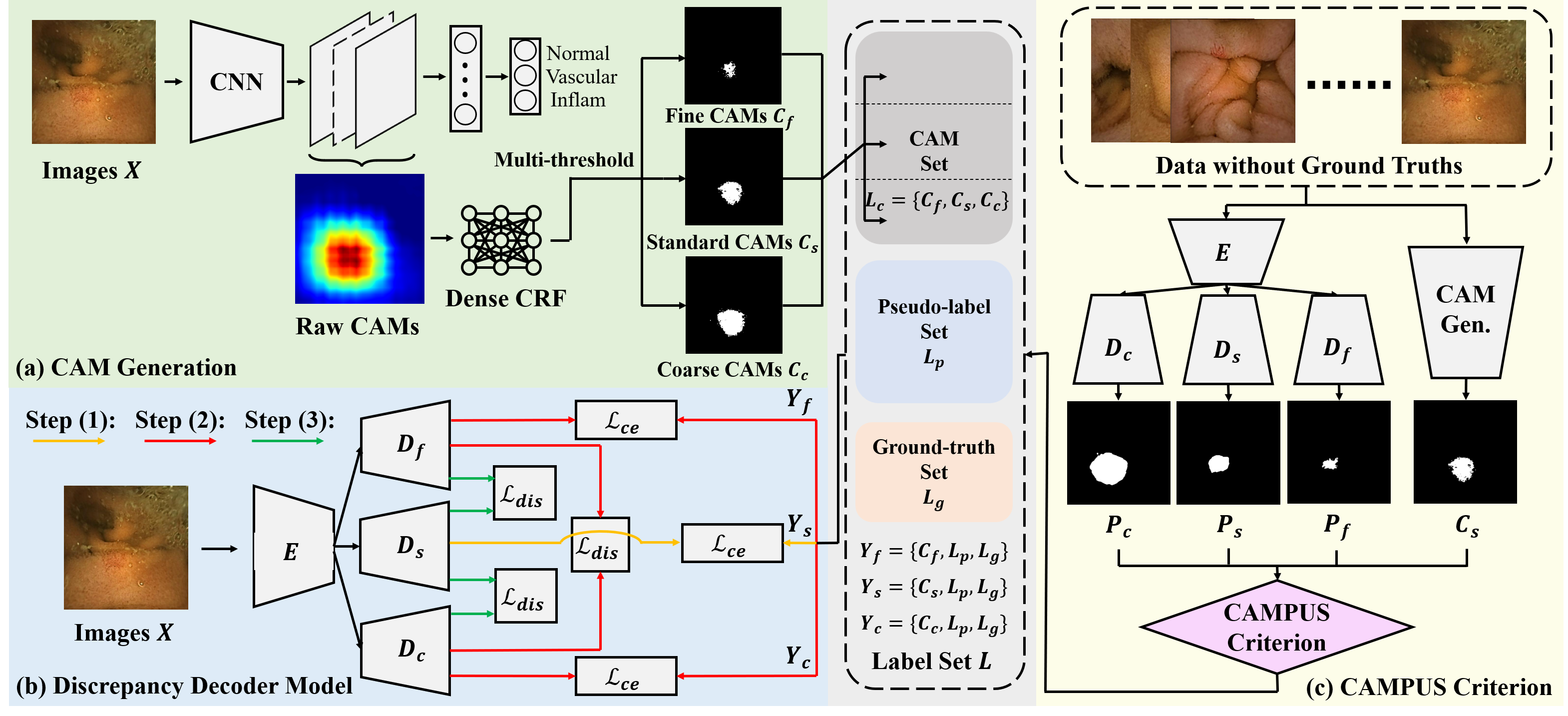}
\caption{The overview of our proposed DEAL method. (a) The CAM generation with three different propensities. (b) The training of discrepancy decoder model with three steps. (c) The CAMPUS criterion based on model divergence and CAM divergence.} 
\label{fig1}
\end{figure}

\section{Methodology}
The overview of our work with three components is demonstrated in Fig.~\ref{fig1}: (a) CAM generation, (b) discrepancy decoder model, (c) CAMPUS criterion. First, we train a classification model based on image-level labels to generate standard CAMs $C_s$, coarse CAMs $C_c$ and fine CAMs $C_f$. For segmentation, there are CAMs in the initial label set $L_c$, and the discrepancy decoder model $M_d$ is trained by $L_c$ under a novel training scheme. In each active learning cycle, guided by DEAL, we select $K_g$ samples for human labeling to form ground-truth set $L_g$ and $K_p$ samples for pseudo labels to form pseudo-label set $L_p$. Since the pseudo labels have no label burden, $K_p$ is an adaptive parameter rather than a hyperparameter. In the end, our label set $L$ consists of CAM set $L_c$, pseudo-label set $L_p$, and ground-truth set $L_g$, which guarantees the performance of the segmentation model $M_d$ with as little labor cost as possible.

\subsection{CAM Generation}
Image segmentation requires a huge amount of pixel-level labels, resulting in labeling burden, especially in medical images which need expert experience. Fortunately, supervised by image-level labels only, the CAMs can roughly locate the foreground regions and provide good seeds for segmentation. In this paper, we follow the main-stream approach to train a fully supervised classification model with image-level labels. We replace the stride with dilation in the last layer to increase the map size on ResNet-50. Meanwhile, we also use regularization terms, drop blocks, and data augmentation to guarantee a good performance. After training, we calculate the CAMs via the Grad-CAM method~\cite{selvaraju2017grad}:
\begin{equation}
    \alpha_{i}^{c} = \frac{1}{Z} \sum_{h=1}^{m} \sum_{w=1}^{n} \frac{\partial S_c}{\partial A_{hw}^{i}}
\end{equation}
\begin{equation}
    L_{Grad-CAM}^{c} = ReLU(\sum_i \alpha_{i}^{c} A^{i})
\end{equation}
where $\alpha_{i}^{c}$ is the weight connecting the $i^{th}$ feature map with the $c^{th}$ class, $S_c$ is the classification score of class $c$, $A_{hw}^{i}$ is the value of $i^{th}$ feature map $h^{th}$ row $w^{th}$ column,  $Z$ is the number of pixels in the feature map ($Z = \sum_h \sum_w 1$).

To deal with the high false positives problem~\cite{qu2020weakly} in medical images, we post process the CAMs with dense CRF~\cite{chen2017deeplab} to refine the boundaries and reduce noise. Nevertheless, we can not eliminate the noise completely. To evaluate the uncertainty of the CAM generation, we leverage multi-threshold to generate standard, coarse and fine CAMs, and prepare to train the discrepancy decoder model. In Fig.~\ref{fig1}(a), we set the best, low and high thresholds to generate the standard, coarse and fine CAMs, respectively. Compared with the standard CAMs, the coarse CAMs have coarser boundaries with higher false positives and the fine CAMs have more detailed textures with higher false negatives. These different CAMs train the discrepancy decoders to form corresponding propensities, described in detail below.

\subsection{Discrepancy Decoder Model}
Intuitively, the straightforward way to assess the CAM uncertainty is to measure the difference between model predictions and CAMs. However, due to the noisy CAMs and the unstable model with noisy training, we can not accurately judge whether this gap is caused by CAM uncertainty or model uncertainty. With the help of multiple discrepancy models, we can estimate model uncertainty and CAM uncertainty, respectively. Different from the model ensemble approach~\cite{2018ensembles}, these discrepancy models should not be arbitrary. They should have two properties: (1) The decision boundary between discrepancy models and the standard model should be adjacent to prevent departure from the standard model; (2) The prediction discrepancy should be positively related to the CAM uncertainty instead of being arbitrary. Next, we will detail the discrepancy decoder model and training scheme.

\subsubsection{Network Architecture:}
Our novel discrepancy decoder model is shown in Fig.~\ref{fig1}(b). Without loss of generality, we directly leverage U-Net~\cite{unet} used in medical image segmentation. For the encoder $E$, we adopt ResNet-18~\cite{he2016resnet} and replace the stride with dilation to increase the receptive field. We use the same structure of three U-Net decoders $D_s$, $D_c$ and $D_f$. For fairness, the model's performance is evaluated by $D_s$, and other decoders only work in label selection. 

\subsubsection{Training Procedure:}
Let $E$ denotes the encoder parameterized by $\theta_E$. The decrepancy decoders $D_s$, $D_c$ and $D_f$ are parameterized by $\theta_{D_s}$, $\theta_{D_c}$ and $\theta_{D_f}$, respectively. For input images $X$, there are label sets $Y_s$, $Y_c$ and $Y_f$ from the standard CAMs $C_s$, coarse CAMs $C_c$ and fine CAMs $C_f$, respectively. After active label selection, we replace some CAMs with pseudo labels or ground truths. In Fig.~\ref{fig1}(b), our novel training scheme consists of three steps: 

(1) We train the encoder $E$ and decoder $D_s$ using $Y_s$. 
The objective is
\begin{equation}
    \min_{\theta_E,\theta_{D_s}}{\mathcal{L}_{ce}(D_s(E(X)),Y_s)},
\end{equation}
where $\mathcal{L}_{ce}$ denotes the loss function with Cross-Entropy and Dice loss.

(2) We duplicate $\theta_{D_s}$ to $\theta_{D_c}$ and $\theta_{D_f}$, fix the encoder $E$, and train the two decoders $D_c$ and $D_f$ with $Y_c$ and $Y_f$ to maximize the prediction discrepancy, which forms different prediction propensities and makes the discrepancy larger. To get a coarse decoder and a fine decoder, the objective is:
\begin{equation}
    \min_{\theta_{D_c},\theta_{D_f}}{\mathcal{L}_{ce}(D_c(E(X)),Y_c)+\mathcal{L}_{ce}(D_f(E(X)),Y_f)-\mathcal{L}_{dis}(D_c(E(X)),D_f(E(X)))},
\end{equation}
where $\mathcal{L}_{dis}$ denotes the L1 distance between the two discrepancy predictions, which is proved effective in~\cite{saito2018maximum}.

(3) We fine-tune the decoders $D_c$ and $D_f$ to minimize the prediction discrepancy with $D_s$, making the boundary of the discrepancy decoders always surround the boundary of the standard decoder. The objective function is defined as:
\begin{equation}
    \min_{\theta_{D_c},\theta_{D_f}}{\mathcal{L}_{dis}(D_c(E(X)),D_s(E(X)))+\mathcal{L}_{dis}(D_f(E(X)),D_s(E(X)))}.
\end{equation}

In the training process, we first train the model according to step (1). Then we iterate steps (2) and (3) for several epochs. After that, we obtain a model with a shared encoder, a standard decoder and two discrepancy decoders.

\subsection{CAMPUS Criterion}
Evaluating the CAM uncertainty and the model uncertainty is crucial for selection. In this regard, we define the model divergence and the CAM divergence for selecting pseudo labels and ground truths in the CAMPUS criterion. To avoid weights that are difficult to balance between different criteria, we do not use additive combinations but better multiplicative combinations.

\subsubsection{Model Divergence:}
The model divergence represents the model uncertainty, including prediction entropy and divergence among three discrepancy decoders. The prediction divergence is related to the standard prediction $P_s$, coarse prediction $P_c$ and fine prediction $P_f$. If one sample has three very different predictions, the prediction is likely to be unreliable. In detail, we measure the score of model divergence $S_{md}$ based on the formula below. 
\begin{equation}
    S_{md} = S_e \cdot(\mathcal{D}(P_s,P_c) + \mathcal{D}(P_s,P_f) + \mathcal{D}(P_c,P_f)),
\end{equation}
where $\mathcal{D} = 1 - \frac{2TP}{FP+2TP+FN}$ defines the Dice distance between predictions. Here TP, FP, TN, and FN are true positive, false positive, true negative and false negative. The prediction entropy $S_e = - \frac{1}{N}\sum_{i=1}^N p_{si}\log p_{si}$ denotes the prediction uncertainty of the standard decoder, where $p_{si}$ is the probability of the $i^{th}$ pixel in $P_s$.

\subsubsection{CAM Divergence:}
The CAM divergence defines the divergence between predictions and CAMs, positively related to the CAM uncertainty. To eliminate the outliers of the prediction, we design the score $S_{cd}$ by the distance between CAMs and two predictions that are closest to the CAMs. 
\begin{equation}
\begin{aligned}
    S_{cd} = sum\{{D}_{3}\} - max\{{D}_{3}\}, \mathcal{D}_{3} = \{\mathcal{D}(P_s,Y_s), \mathcal{D}(P_c,Y_s), \mathcal{D}(P_f,Y_s)\}
\end{aligned}
\end{equation}

In pseudo-label selection, we should pick out the samples with small model divergence and large CAM divergence for pseudo labels, implying noisy CAMs and accurate predictions. The criterion $S_p$ is 
\begin{equation}
    S_p = (3 - S_{md}) \cdot S_{cd}.
\end{equation}
We sort the images in increasing order, locate the knee point $K_p$ of the score curve by KneeLocator~\cite{satopaa2011finding}, and select the higher point. 

In ground-truth selection, we select uncertainty samples with large model divergence and CAM divergence for human labeling. The criterion $S_g$ is
\begin{equation}
    S_g = S_{md}\cdot S_{cd}.
\end{equation}
We sort $S_g$ in increasing order and select top $K_g$ values to annotate manually.

\begin{table}[tp!]
\caption{The segmentation performance of the competing approaches. The 95\% of the performance with 100\% ground truths is 0.7840, which is a line for achieving comparable performance to fully supervised learning in active learning\cite{2020ViewAL}.}
\label{tab1}
\begin{tabular}{c|ccccc}
\hline
\multirow{2}{*}{Method}                                            & \multicolumn{5}{c}{Dice with Ground Truths}                                                                                                                                                                                                                                                                                                                                                                                                            \\ \cline{2-6} 
                                                                   & \multicolumn{1}{c|}{0\%}                                                                        & \multicolumn{1}{c|}{10\%}                                                               & \multicolumn{1}{c|}{20\%}                                                               & \multicolumn{1}{c|}{30\%}                                                               & 100\%                                                                      \\ \hline
Random                                                             & \multicolumn{1}{c|}{\multirow{8}{*}{\begin{tabular}[c]{@{}c@{}}0.7364\\ (0.0254)\end{tabular}}} & \multicolumn{1}{c|}{0.7598(0.0250)}                                                     & \multicolumn{1}{c|}{0.7836(0.0214)}                                                     & \multicolumn{1}{c|}{0.7837(0.0223)}                                                     & \multirow{9}{*}{\begin{tabular}[c]{@{}c@{}}0.8253\\ (0.0163)\end{tabular}} \\ \cline{1-1} \cline{3-5}
Dice                                                               & \multicolumn{1}{c|}{}                                                                           & \multicolumn{1}{c|}{0.7645(0.0274)}                                                     & \multicolumn{1}{c|}{0.7871(0.0253)}                                                     & \multicolumn{1}{c|}{0.7933(0.0187)}                                                     &                                                                            \\ \cline{1-1} \cline{3-5}
VAAL~\cite{sinha2019vaal}                                                               & \multicolumn{1}{c|}{}                                                                           & \multicolumn{1}{c|}{0.7644(0.0252)}                                                     & \multicolumn{1}{c|}{0.7810(0.0248)}                                                     & \multicolumn{1}{c|}{0.7973(0.0171)}                                                     &                                                                            \\ \cline{1-1} \cline{3-5}
CoreSet~\cite{sener2017coreset}                                                    & \multicolumn{1}{c|}{}                                                                           & \multicolumn{1}{c|}{0.7598(0.0245)}                                                     & \multicolumn{1}{c|}{0.7762(0.0279)}                                                     & \multicolumn{1}{c|}{0.7937(0.0202)}                                                     &                                                                            \\ \cline{1-1} \cline{3-5}
CoreGCN~\cite{caramalau2021gcn}                                                  & \multicolumn{1}{c|}{}                                                                           & \multicolumn{1}{c|}{0.7697(0.0215)}                                                     & \multicolumn{1}{c|}{0.7845(0.0167)}                                                     & \multicolumn{1}{c|}{0.7899(0.0251)}                                                     &                                                                            \\ \cline{1-1} \cline{3-5}
UncertaintyGCN~\cite{caramalau2021gcn}                     & \multicolumn{1}{c|}{}                                                                           & \multicolumn{1}{c|}{0.7598(0.0233)}                                                     & \multicolumn{1}{c|}{0.7796(0.0266)}                                                     & \multicolumn{1}{c|}{0.7953(0.0225)}                                                     &                                                                            \\ \cline{1-1} \cline{3-5}
GGS~\cite{dai2020ggs}                                                           & \multicolumn{1}{c|}{}                                                                           & \multicolumn{1}{c|}{0.7618(0.0163)}                                                     & \multicolumn{1}{c|}{0.7836(0.0204)}                                                     & \multicolumn{1}{c|}{0.7988(0.0227)}                                                     &                                                                            \\ \cline{1-1} \cline{3-5}
\begin{tabular}[c]{@{}c@{}}DEAL\\ (w/o pseudo labels)\end{tabular} & \multicolumn{1}{c|}{}                                                                           & \multicolumn{1}{c|}{0.7735(0.0242)}                                                     & \multicolumn{1}{c|}{0.7967(0.0134)}                                                     & \multicolumn{1}{c|}{0.8066(0.0172)}                                                     &                                                                            \\ \cline{1-5}
DEAL                                                              & \multicolumn{1}{c|}{\textbf{\begin{tabular}[c]{@{}c@{}}0.7626\\ (0.0270)\end{tabular}}}         & \multicolumn{1}{c|}{\textbf{\begin{tabular}[c]{@{}c@{}}0.7947\\ (0.0211)\end{tabular}}} & \multicolumn{1}{c|}{\textbf{\begin{tabular}[c]{@{}c@{}}0.7973\\ (0.0149)\end{tabular}}} & \multicolumn{1}{c|}{\textbf{\begin{tabular}[c]{@{}c@{}}0.8083\\ (0.0215)\end{tabular}}} &                                                                            \\ \hline
\end{tabular}
\end{table}

\section{Experiments and Results}
\subsection{Experimental Settings}
\subsubsection{Datasets:}
We conducted experiments on the CAD-CAP WCE dataset \cite{dray2018cad}. It contains 1812 images with both image-level and pixel-level labels available (600 normal images, 605 vascular images and 607 inflammatory images). In data preprocessing, we adopted the deformation field~\cite{guo2020semi} to preprocess all images to $320\times320$. We split labeled data by 5-fold cross-validation and calculated the $Dice = \frac{2TP}{FP+2TP+FN}$ as the metric to evaluate the performance.

\begin{table}[tp!]
\centering
\caption{The ablation studies for our proposed method. We show the changes compared to the initial under different settings in  pseudo-label and 20\% ground-truth selection.}
\label{tab2}
\begin{tabular}{ccc|cc}
\hline
\multicolumn{1}{c|}{\begin{tabular}[c]{@{}c@{}}Discrepancy \\ Model\end{tabular}} & \multicolumn{1}{c|}{\begin{tabular}[c]{@{}c@{}}Model \\ Divergence\end{tabular}} & \begin{tabular}[c]{@{}c@{}}CAM \\ Divergence\end{tabular} & \multicolumn{1}{c|}{\begin{tabular}[c]{@{}c@{}}Pseudo-label \\ Selection\end{tabular}} & \begin{tabular}[c]{@{}c@{}}Ground-truth\\ Selection\end{tabular} \\ \hline
$\times$        & $\times$          & \checkmark    & -8.70\% & +3.77\%  \\
\checkmark      & \checkmark        & $\times$      & +2.01\% & +3.51\%   \\
\checkmark      & $\times$          & \checkmark    & +2.07\% & +4.42\%   \\
\checkmark      & \checkmark        & \checkmark    & +2.62\% & +6.09\%   \\ \hline
\end{tabular}
\end{table}

\subsubsection{Competing Approaches:}
For label selection, we compared our DEAL with various types of active learning methods: (1) Random; (2) Dice, which computes the CAM uncertainty naively using the Dice between predictions and CAMs; (3) VAAL~\cite{sinha2019vaal} (diversity-based); (4) CoreSet~\cite{sener2017coreset} (representativeness-based); (5) CoreGCN~\cite{caramalau2021gcn} (representativeness-based); (6) UncertaintyGCN~\cite{caramalau2021gcn} (uncertainty-based); (7) GGS~\cite{dai2020ggs} (gradient-based). (8) Our DEAL without pseudo labels; (9) Our DEAL. Since methods (3)-(7) require an initial label set as a reference, we randomly initialized half of the budget and actively selected the other half in the first cycle. From the second cycle, all methods selected labels normally to reach the budget.

\subsubsection{Training Setup:}
We conducted our experiment on a single GTX 3090 GPU with Pytorch. In CAM generation, we trained a ResNet-50 model to classify images into normal, vascular and inflammatory to generate high-quality CAMs. The standard, coarse and fine CAMs about vascular were generated by thresholds 0.8, 0.75, and 0.85 as muti-threshold masks for segmentation. In segmentation, we trained our discrepancy decoder model to segment the bleeding and background with the Adam optimizer of learning rate 0.003 for 50 epochs. To eliminate the effects of the model ensemble and training process, we only used the output of the standard decoder as the model's predictions in the test. We conducted all segmentation experiments under fixed random seeds and 5-fold cross-validation and calculated the mean and variance.

\subsection{Results}
\subsubsection{Evaluation of Label Selection:}
After training with CAMs as the initial, all methods selected 10\% CAMs actively replaced by ground truths via human labeling in each cycle. Three rounds of active selection were performed. Our DEAL selected pseudo labels before the first cycle because of the CAMPUS criterion. Tabel~\ref{tab1} shows the selection performance of the competing approaches. We can see that our DEAL (w/o pseudo labels) and DEAL outperform other methods on 10\%, 20\%, and 30\% ground truths, respectively. By comparing DEAL and DEAL (w/o pseudo labels), we concluded labeling gain of pseudo-label selection is significant. After pseudo-label selection, our DEAL is 2.6\% higher than the initial model without any ground truth. We surprisingly find that our DEAL outperforms 95\% performance of full supervised training with only 10\% ground truths, which is much better than other methods. Our DEAL can reach performance saturation quickly under a few ground truths and achieve 0.8083 on Dice finally. Obviously, other methods are far inferior to ours. Our DEAL saves 90\% ground truths and achieves the comparable performance of full supervised training, which is of great significance to medical image segmentation.

\begin{figure}[tp!]
\centering
\includegraphics[width=0.9\textwidth]{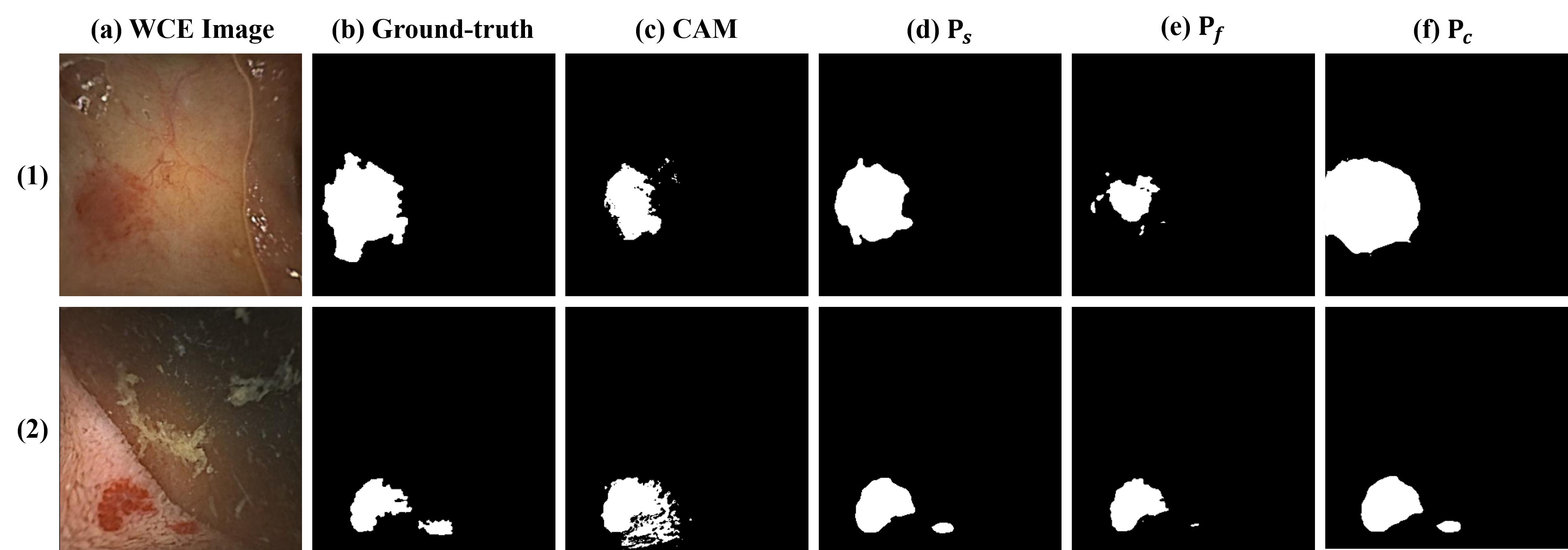}
\caption{(1) In ground-truth selection, we pick out samples with large divergence in three predictions and the CAM. (2) In pseudo-label selection, we select samples with small model divergence and large CAM divergence.} 
\label{fig2}
\end{figure}

\subsubsection{Ablation Studies:}
We conducted ablation studies on the discrepancy decoder model, model divergence, and CAM divergence, respectively. Specifically, we compared the pseudo-label and 20\% ground-truth selection performance under different settings and showed the performance changes compared with the initial model. Note that the none-discrepancy decoder model can not obtain model divergence. Table~\ref{tab2} shows the effect of each part of DEAL on model performance. We find the single network without model divergence evaluation can not accurately select the pseudo labels, resulting in  -8.80\% performance degradation. In the CAMPUS criterion, both the model divergence and the CAM divergence play a crucial role, and lacking any of them is massive destruction of performance.

\subsubsection{Qualitative Analysis:}
To demonstrate our DEAL evaluates the CAM uncertainty more intuitively, we visualized the WCE images, ground truths, CAMs, and predictions from the discrepancy decoder model on Fig.~\ref{fig2}, respectively.

Fig.~\ref{fig2} shows our discrepancy decoder model can generate standard, coarse and fine predictions, respectively. In pseudo-label selection, the predictions between three discrepancy decoders are similar but very different from CAMs. In ground-truth selection, both three predictions and CAMs have large gaps which are positively correlated with the gaps between CAMs and ground truths.

\section{Conclusions}
In this paper, we propose a novel Discrepancy-basEd Active Learning (DEAL) approach to bridge the gaps between the CAMs and the ground truths with pseudo labels and a few annotations. With the designed discrepancy decoder model and CAMPUS criterion, our approach can detect the samples with noisy CAMs and replace them with high-quality pseudo labels and ground truths to minimize human labeling costs. As the first label selection method to advance weakly supervised to approach fully supervised performance, our DEAL saves labeling burden significantly. In the future, our method can be transferred to a wide variety of medical image applications and save the labeling burden.

\subsubsection{Acknowledgements.} The work described in this paper was supported by National Key R\&D program of China with Grant No. 2019YFB1312400, Hong Kong RGC CRF grant C4063-18G, and Hong Kong RGC GRF grant \# 14211420.

\bibliographystyle{splncs04}
\bibliography{mybib}

\end{document}